\documentclass[conference]{IEEEtran}
\pdfoutput=1
\usepackage{cite}

\ifCLASSINFOpdf
   \usepackage[pdftex]{graphicx}
  \DeclareGraphicsExtensions{.pdf,.jpeg,.png}
\else
   \usepackage[dvips]{graphicx}
  \DeclareGraphicsExtensions{.eps}
\fi
\usepackage{amsmath}
\ifCLASSOPTIONcompsoc
  \usepackage[caption=false,font=normalsize,labelfont=sf,textfont=sf]{subfig}
\else
  \usepackage[caption=false,font=footnotesize]{subfig}
\fi
\usepackage{url}

\usepackage[gen]{eurosym}

\usepackage{multirow}

\usepackage{enumitem}

\usepackage{amssymb}

\usepackage{soul}
\usepackage{tikz}
\usetikzlibrary{calc}
\makeatletter
\newif\if@anonymize

\@anonymizefalse  

\if@anonymize
  \newcommand{\highlight@DoHighlight}{
    \fill [outer sep = -15pt, inner sep = 0pt, color=black]
          ($(begin highlight)+(0,8pt)$) rectangle ($(end highlight)+(0,-3pt)$) ;
  }

  \newcommand{\highlight@BeginHighlight}{
    \coordinate (begin highlight) at (0,0) ;
  }

  \newcommand{\highlight@EndHighlight}{
    \coordinate (end highlight) at (0,0) ;
  }

  \newdimen\highlight@previous
  \newdimen\highlight@current
  \newlength{\item@width}

  \DeclareRobustCommand*\anonymize{%
    \SOUL@setup
    \def\SOUL@preamble{%
      \begin{tikzpicture}[overlay, remember picture]
        \highlight@BeginHighlight
        \highlight@EndHighlight
      \end{tikzpicture}%
    }%
    \def\SOUL@postamble{%
      \begin{tikzpicture}[overlay, remember picture]
        \highlight@EndHighlight
        \highlight@DoHighlight
      \end{tikzpicture}%
    }%
    \def\SOUL@everyhyphen{%
      \discretionary{%
        \SOUL@setkern\SOUL@hyphkern
        \SOUL@sethyphenchar
        \tikz[overlay, remember picture] \highlight@EndHighlight ;%
      }{%
      }{%
        \SOUL@setkern\SOUL@charkern
      }%
    }%
    \def\SOUL@everyexhyphen##1{%
      \SOUL@setkern\SOUL@hyphkern
      \settowidth{\item@width}{##1}%
      \makebox[\item@width]{}%
      \discretionary{%
        \tikz[overlay, remember picture] \highlight@EndHighlight ;%
      }{%
      }{%
        \SOUL@setkern\SOUL@charkern
      }%
    }%
    \def\SOUL@everysyllable{%
      \begin{tikzpicture}[overlay, remember picture]
        \path let \p0 = (begin highlight), \p1 = (0,0) in \pgfextra
          \global\highlight@previous=\y0
          \global\highlight@current =\y1
        \endpgfextra (0,0) ;
        \ifdim\highlight@current < \highlight@previous
          \highlight@DoHighlight
          \highlight@BeginHighlight
        \fi
      \end{tikzpicture}%
      \settowidth{\item@width}{\the\SOUL@syllable}%
      \makebox[\item@width]{}%
      \tikz[overlay, remember picture] \highlight@EndHighlight ;%
    }%
    \SOUL@
  }
\else
  \newcommand{\anonymize}[1]{#1}
\fi
\makeatother

\begin{document}
%
\title{e-Counterfeit: a mobile-server platform for document counterfeit detection}

\author{\IEEEauthorblockN{\anonymize{Albert Berenguel}\IEEEauthorrefmark{1}\IEEEauthorrefmark{2}\anonymize{,
Oriol Ramos Terrades}\IEEEauthorrefmark{1}\anonymize{,
Josep Llad\'os}\IEEEauthorrefmark{1} \anonymize{and
Cristina Ca\~nero}\IEEEauthorrefmark{2}}
\IEEEauthorblockA{\IEEEauthorrefmark{1}\anonymize{Computer Vision Center, Universitat Autonoma of Barcelona, Barcelona, Spain} \\
\anonymize{Email: \{aberenguel,oriolrt,josep\}@cvc.uab.es}}
\IEEEauthorblockA{\IEEEauthorrefmark{2}\anonymize{ICAR Vision Systems S.A, Barcelona, Spain} \\
\anonymize{Email: cristina@icarvision.com}}
}


%


\maketitle

\begin{abstract} 
This paper presents a novel application to detect counterfeit identity documents forged by a scan-printing operation. Texture analysis approaches are proposed to extract validation features from security background that is usually printed in documents as IDs or banknotes. The main contribution of this work is the end-to-end mobile-server architecture, which provides a service for non-expert users and therefore can be used in several scenarios. The system also provides a crowdsourcing mode so labeled images can be gathered, generating databases for incremental training of the algorithms.
\end{abstract}


%
\IEEEpeerreviewmaketitle

\section{Introduction}

Making a fake passport is easy, making a good fake passport is very, very hard. Probably there are few criminal organizations in the world which can produce a counterfeit visa or passport good enough to fool professional passport control. Counterfeit detection has traditionally been a task for law enforcement agencies. EUROPOL and INTERPOL central offices are combating document and banknote counterfeiting~\cite{interpol,europol}. They have destined millions of euros in funds to provides technical databases, forensic support, training and operational assistance to its member countries. Despite all these efforts counterfeit detection remains an open issue. There are many different strategies used to fake an identity document (ID) like the alteration of a real passport, impersonation of the legitimate owner or printing false information on a stolen blank real paper, to cite some. An important security feature that serves against counterfeiting and manipulation of documents is the background/security printing. The security printing can be classified into two categories: 

\begin{enumerate}
   \item Security printing techniques and printing processes used: microprinting as well as security inks (e.g. optical variable ink, ultraviolet or infrared ink). This security features can only be checked with specialized equipment like ultraviolet lamp or a magnified glass, therefore are not possible to check if our target of image acquisition is the general smartphones cameras.
   \item Print designs and security elements: guilloches or fine-line patterns, rainbow coloring, etc., discernible to the naked eye. We focus in this category of elements.
\end{enumerate}

The counterfeiter requires high technical specialized printing equipment to reproduce this background printing techniques. Having this equipment is not feasible for the majority of the counterfeiters due economic and restricted availability issues. Thanks to this, a large part of counterfeits just follows the procedure of scan a real document, alter the data and then print the document with a common commercial printer. We refer to this methodology of counterfeit document generation as \textit{scan-printing} procedure. Following the scan-printing procedure, we expect that the background print design will loose detail hence it will be possible to classify it as counterfeit. We point our efforts to detect this kind of low quality counterfeit documents. It is important to note that most of the print design methodology that is possible to find in the security documents, were first created and used to detect counterfeit money banknotes. 
The system proposed in this work has been designed for ID documents, however, as mentioned before, it is generic enough so it can be used for other documents with security background texture as banknotes. The main functionality is the analysis of the document  authenticity from a single image using a mobile phone camera within a non-controlled environment. This process is used by services or products that require a genuine identification of the client, such as renting a car, opening a bank account, applying for a loan, checking-in in an hotel, etc. Once authenticity of the document is validated, the purpose of the platform where the service is integrated can take place (reading and storing the personal information of the document holder).

The main contribution is a mobile-server framework to detect counterfeit documents. This end-to-end system provides: a low cost solution fit to be used with common smartphones, flexibility for further updates and robust validation methods. The mobile-server framework also intends to address the lack of tools to generate datasets acquired from smartphone devices. 
We have published the framework code to help future researchers\footnote{\label{server_github}\url{https://github.com/gitabcworld/e-Counterfeit}.}. 

The remainder of the paper is organized as follows. In section \ref{section:related_work}, we review the existing frameworks for banknote counterfeit detection. Section \ref{section:system_architecture} describes in detail each one of the components of the end-to-end framework proposed and \ref{section:counterfeit_module} explains in detail the Counterfeit module. In section \ref{section:dataset} the dataset is presented. At section \ref{section:experiments} we explain the current set-up and the discussion of the results. Finally, in section \ref{section:conclusions} we draw some conclusions and outline rewarding avenues for future work.

\begin{figure*}[h!]
	\centering
    \captionsetup{justification=centering}
  	\includegraphics[width=\textwidth,height=5cm]{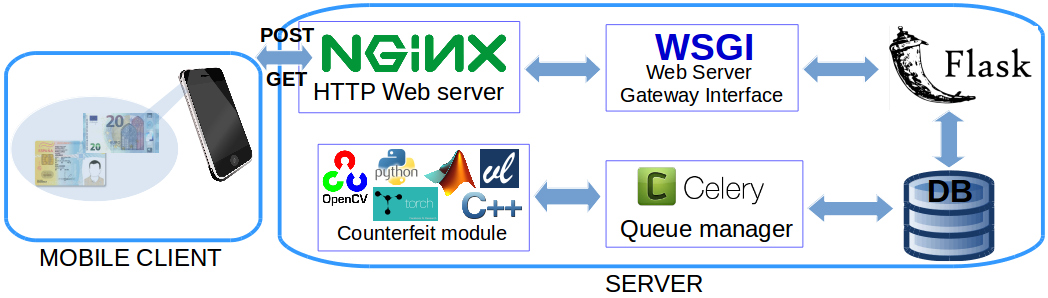}
  	\caption{Server-client counterfeit framework. A document is acquired with a mobile application by the user. The image is sent to the server which stores this image into a database through the web-server. Afterwards a queue manager manages several workers, which are reading simultaneously from the database to process the images with the Counterfeit module. Finally the results are sent back to the user. The WSGI module is used to connect the HTTP messages with Python Flask module that centralizes all the operations.}
    \label{server_app}
\end{figure*}

\section{Related work}\label{section:related_work}

Several private companies are specialized in ID document counterfeit using smartphone applications~\cite{Icarvision,KeesingTechnologies,Jumio,IDscan}. However there is few academic research published about this topic. The authors in~\cite{DocScope} developed an operational pilot project to be used in mobile context focused on document background and photo area printing. On the other hand, the counterfeit analysis field applied to banknotes has been a topic of interest for researchers the last years, being this field applicable to ID counterfeiting. Other researchers have focused in a banknote validation method which uses radio frequency identification (RFID) and an NFC-enabled smartphone~\cite{eldefrawy2015banknote}. Although this method is computationally less expensive than other methods not all the documents and banknotes have the RFID chips. For this reason, we focus on texture background like the authors in~\cite{lohweg2013banknote}, where they use small patches of the banknote focused in the Intaglio regions instead. A different approach is treat the problem of banknote counterfeit detection as a dictionary problem to cover all possible background texture variations and encode the representation as sparse coding features~\cite{berenguel2016banknote}. They validate the complete document from a single image. The Intaglio zones have been also validated by the authors in~\cite{dewaele2016forensic,pfeifer2016detection}, using Discrete Fourier Transformation (DFT) and models to describing the edge transition respectively. 
They acquire their dataset with a minimum of 1100 dpi within a controlled environment.

\section{System architecture and components}\label{section:system_architecture}

We follow the document acquisition approach given by the authors in~\cite{berenguel2016banknote} and we build a service-oriented architecture (\textbf{SOA}) end-to-end counterfeit validation framework. The proposed SOA application is composed by integration of distributed, separately-maintained and deployed software components. The end-to-end system developed is scalable and provide fast responses. The image acquisition procedure is easy and friendly for the user. Finally, this framework provides a way to store and manage the new data that is being sent to the server for further improvements.

\subsection{Server Framework}\label{section:server_framework}
 
We have built a server which communicates with a client mobile application, see Fig. \ref{server_app}. First a user acquires a document photo which sends through JSON REST API to our server. Through post/get messages we establish a handshake protocol to send the validation image and receive the response information. A web server is set up on top of the operating system to send the HTTP requests, but it could also serve static files like images, JavaScript files, HTML pages, etc. We use NGINX as our web server for its resource efficiency and responsiveness under load. Once the web server has the data, it process the JSON message with Flask, which is a micro-framework for Python focused at web application code. Since a web server cannot communicate directly with Flask, we implement a Web Server Gateway Interface (WSGI) to act as a proxy between the server and Python/Flask. As a summary we have HTTP requests routed from the web server to Flask, which Flask handles appropriately. The responses are then sent right back to the web server and, lastly, back to the end user application.
The Flask module will unpack the JSON message containing the image to process. The image along with other data will be stored in a MySQL database. An ID is assigned to this image and is given back to the client. The image will not be processed immediately, it will be managed by Celery, a queue manager which will be constantly querying through multiple workers the database for new images to process. Meanwhile the mobile clients will establish a handshake protocol intermittently asking for the result to the server using the given ID. Finally each one of the queue manager workers will access the counterfeit module which will process the image and store the result in the database. The counterfeit module integrates into this framework languages as Matlab, OpenCV, VlFeat, C++, Python and Torch, with the purpose to evaluate different existing algorithms written in their respective languages by their authors. Identity documents are also preprocessed to find blurring and highlights although these algorithms are out of the scope of this paper.

\subsection{Mobile client}\label{section:mobile_client}

\begin{figure}
\centering
\subfloat[Adjust photo]{\includegraphics[width=1.7in]{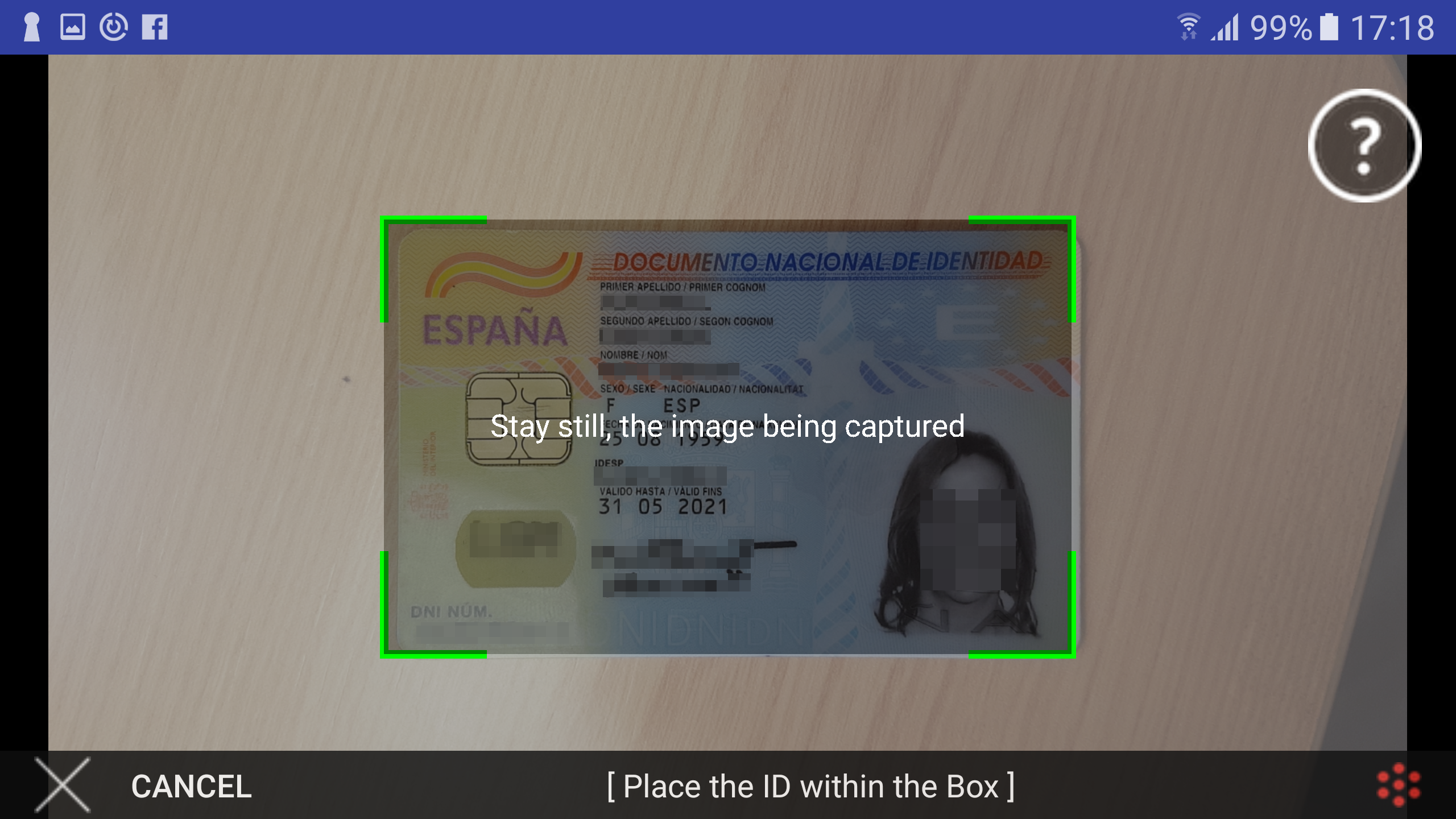}\label{Adjust_photo}} 
\hfill
\subfloat[Crop]{\includegraphics[width=1.7in]{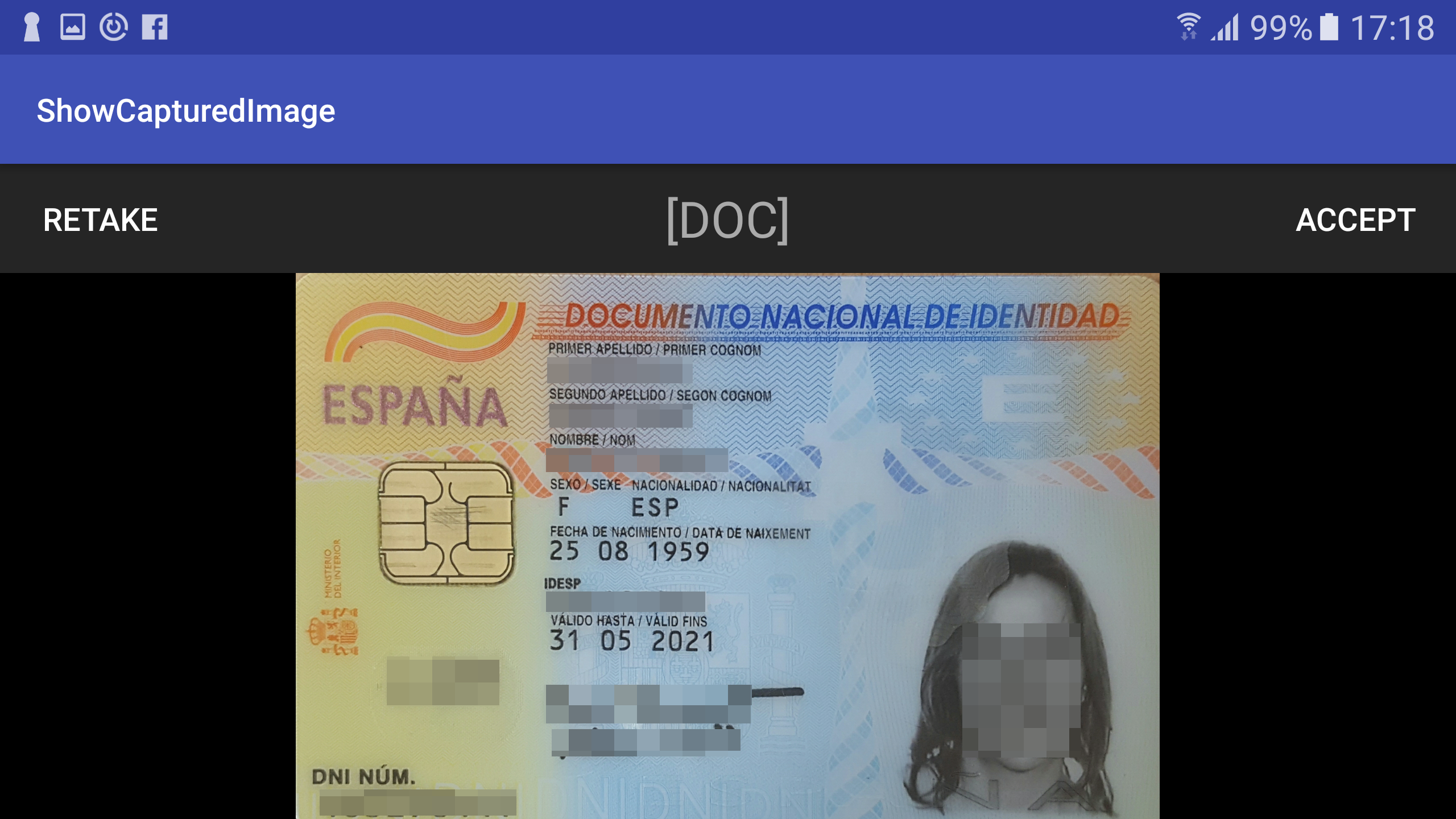}\label{Crop}} 
\hfill
\subfloat[Show results]{\includegraphics[width=\columnwidth]{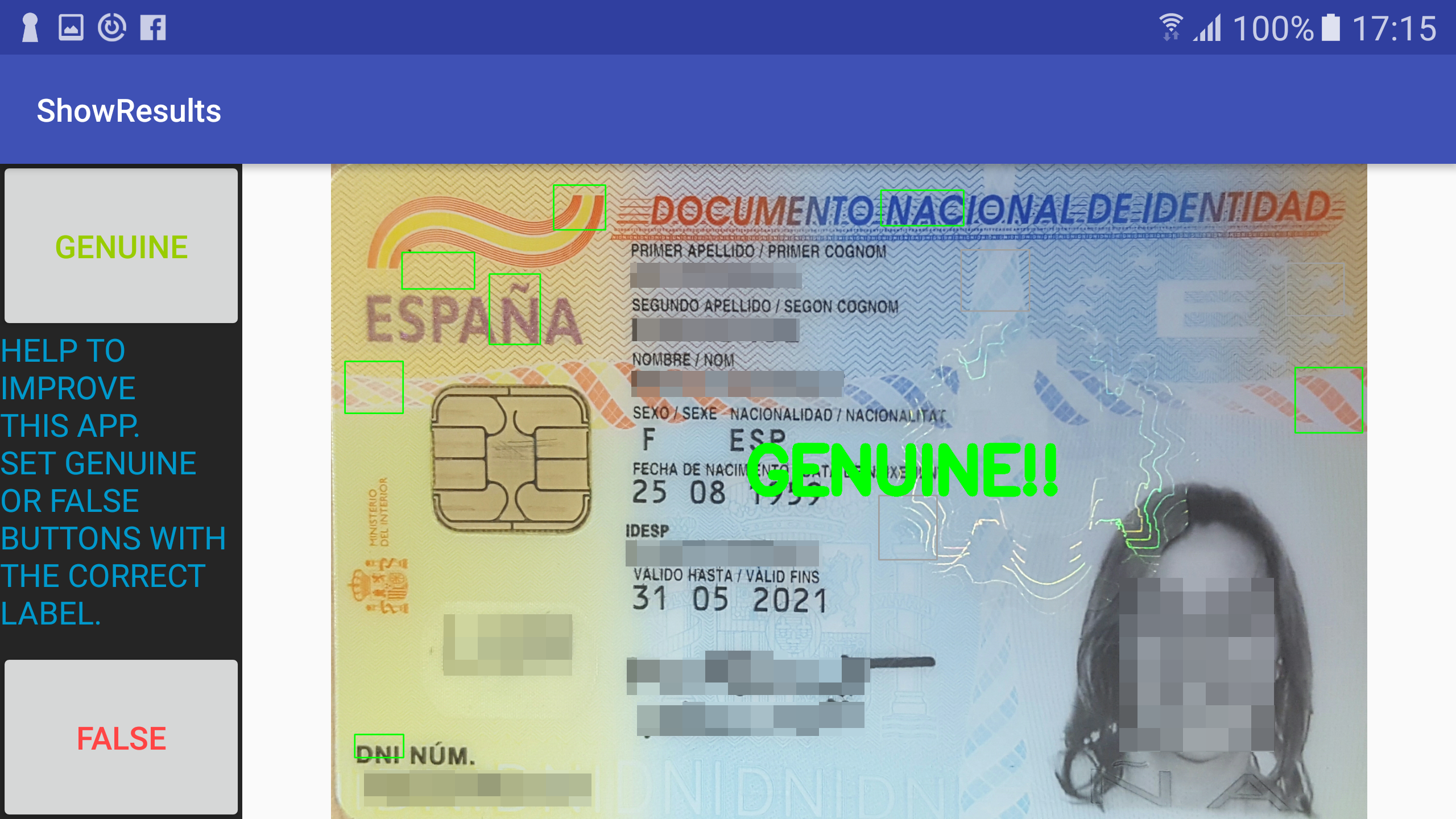}\label{show_results}}
\caption{Screen shots from the mobile application. In \ref{Adjust_photo} the application helps the user to acquire the auto-photo with visual guides. In \ref{Crop} a GrabCut algorithm is applied for a better crop of the margins of the document. Finally a in \ref{show_results} shows the validation of each one of the ROIs and the final decision (genuine or counterfeit) is overlayed. It also ask the user to collaborate with the groundtruth dataset. Better viewed in color. }
\label{client_app}
\end{figure}

The mobile client application is designed to aid a non-expert user through the steps of acquiring a valid photo, see Fig. \ref{client_app}. The first step helps the user guiding visually to fit the document around a rectangle, which adapts its size to the model of the document being acquired. The photo is automatically acquired after checking the document fits the visual guides and the camera is focused. Detecting perfectly the rectangle surrounding the document is a complicated task because of the non-controlled environment acquisition. Clutter, blurring and illumination can affect the precise cropping of the document. 

\begin{figure}
	\centering
    \captionsetup{justification=centering}
  	\includegraphics[width=\columnwidth,height=7cm]{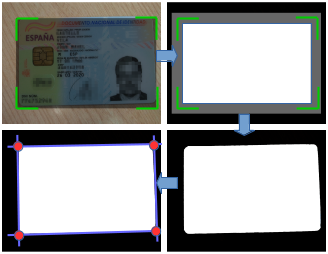}
  	\caption{The GrabCut algorithm and dewarping done in the mobile application. Top-left shows the visual guides to help the photo acquisition. At the top-right image the zones which are foreground (white, interior of visual guides), probably foreground (gray, interior and exterior of visual guides with a margin) and background (black). Bottom-right is the resulting cropping from GrabCut. At bottom-left we find the intersections of the longest lines to perform a dewarping operation to finally correct the perspective of the cropped document.}
    \label{grabcut}
\end{figure}

In the second step we follow the pipeline in Fig. \ref{grabcut} to crop the document. We apply the GrabCut algorithm driven by the position of the visual guides~\cite{boykov2006graph}. We define the foreground as a zone inside the visual guides. The probable foreground will be a zone with a certain margin around the visual guides and the rest will be considered as background. In this operation the image is resized to a 15\% of its size to speed-up the operations. We then binarize the image with the zones labeled as probable foreground returned by the GrabCut algorithm. To find the corners we calculate the contours of the binary mask and filter the lines which do not fulfill a minimum length. Afterwards we calculate all the possible intersection points of those lines, and select only four intersections which are closer to the borders of the image. All this methodology could be replaced by an automatic detection of the rectangle of the document of the document. However we have found that providing a user friendly interface with a visual guide ensures a minimum size, focus and correct cropping of the document to guarantee a correct further processing. Finally the dewarped document is sent to the server and wait asynchronously for its response. 
The last step of the client is to show the results. We overlay on the cropped documents the regions that we are processing and the results of the validation for each one of the regions of interest (ROIs), we also show the final decision for the document if its genuine or counterfeit. Before closing the current view, we ask the users to help for further development and contribute with our dataset labeling the current document if it is genuine or counterfeit. This label is sent to the server and the database is updated accordingly. 
The crowdsourcing task to generate ground truth can be disabled by the user in the configuration. 

\section{Counterfeit module}\label{section:counterfeit_module}
\begin{figure}
	\centering
    \captionsetup{justification=centering}
    \includegraphics[width=\columnwidth,height=10cm]{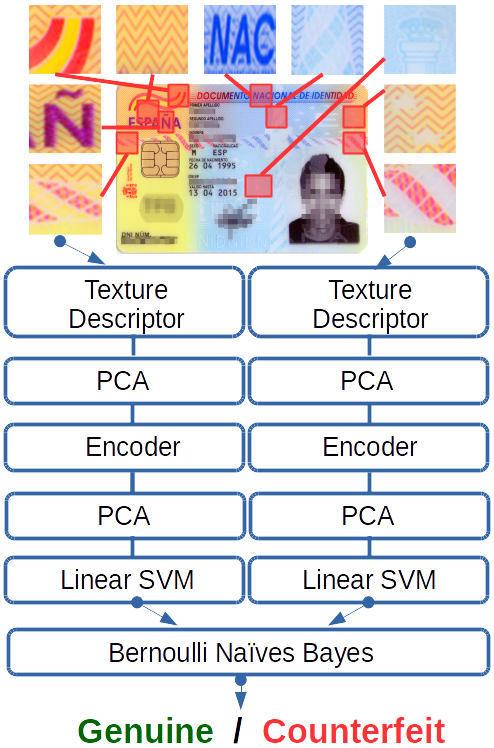}
  	\caption{Proposed architecture for document authentication. The texture descriptor is first reduced with a PCA. After an encoder is learned to improve the feature representation. The encoded texture descriptor dimensionality is again reduced with a PCA to classify each ROI with a linear SVM. A multivariate Bernoulli model predicts the final document decision authenticity.}
    \label{fig:ecounterfeit_features}
\end{figure}

The back-end of the SOA architecture consists in the main module of the service. It is an interchangeable Counterfeit module which validates texture descriptors. 
It is also directly connected with the database which makes possible to train new models automatically as soon as enough images of a certain model are stored. We follow the idea presented by the authors in~\cite{cimpoi2016deep} and build a similar architecture for feature extraction, see Fig. \ref{fig:ecounterfeit_features}. This architecture consists of the following tasks: Texture descriptor extraction, Principal component analysis (PCA), a pooling encoder to improve the representation of the descriptors and a linear Support Vector Machine classifier (SVM). We also include an additional layer that is the Bernoulli Na\"{i}ves Bayes to construct the final decision of labeling a document as genuine or counterfeit. Let us further describe these tasks:

\subsubsection{Texture descriptor} Although any texture feature can be used to represent the textures, we have selected \textbf{dense SIFT} because it is generally very
competitive, outperforming specialized texture descriptors~\cite{cimpoi2016deep,berenguel2016banknote}.  

\subsubsection{Encoder}\label{subsubsection:encoder} A pooling encoder converts the local descriptors to a single feature vector suitable for tasks such as classification with an SVM. We evaluate orderless and order-sensitive pooling encoders. The orderless encoder is invariant to permutations of the input meanwhile the order-sensitive is not. Order-sensitive encoder may be ineffective or even counter-productive in natural texture recognition, but on this counterfeit context it can be helpful to recognize different textured objects and a global description of the texture scene at each ROI. 

The best-known orderless encoder is the Bag of Visual Words (\textbf{BoVW}), which characterizes the distribution of textons~\cite{leung2001representing}. Similarly to BoVW, Vector of Locally-Aggregated Descriptors (\textbf{VLAD})  and Fisher vector (\textbf{FV}), assigns local descriptor to elements in a visual dictionary obtained with $K$-means and Gaussian Mixture Models (GMM) respectively~\cite{jegou2010aggregating,perronnin2007fisher}. BoVW only stores visual words occurrences, meanwhile VLAD accumulates first-order descriptor statistics and FV uses both first and second order statistics of the local image descriptors. As a generalization of $K$-means, we also include as orderless encoder the $\textbf{K}$\textbf{-SVD} dictionary learning algorithm, which creates a dictionary for sparse representations via a singular value decomposition approach~\cite{aharon2006rm}.

On the other hand, Spatial pyramid pooling (SPP) is the most common order-sensitive encoder method~\cite{lazebnik2006beyond}. It divides the image in subregions, computes any encoder for this regions and afterwards stacks the results. We use the spatial pyramid histogram representation \textbf{ScSPM} where the encoded descriptor is the concatenation of local histograms in various partitions of different scales~\cite{yang2009linear}. 
\subsubsection{\textbf{PCA}} The $128$-dimensional descriptors extracted from the texture descriptors step are reduced using PCA. Besides improving the classification accuracy, this significantly reduces the size of the posterior encoding dimensionality. We also include a PCA after the encoder, because VLAD and FV are usually highly compressible vectors so we further reduce the descriptor encoding for comparison purposes~\cite{parkhi2014compact}. 
\subsubsection{\textbf{Linear SVM}} The learning uses a standard nonlinear SVM solver. We train a specific classifier for each ROI for every document. At this point we predict a genuine or counterfeit binary label value determining the ROI authenticity. We normalize the texture descriptors encodings to zero mean and unit variance before SVM classifier. 
\subsubsection{\textbf{Bernoulli Na\"{i}ves Bayes}} From the previous step we obtain a binary feature vector for the ROIs, however we need the final document decision. 
We learn a na{\"\i}ve Bayes classifier according to multivariate Bernoulli distributions, where the decision rule is based on Eq.~\eqref{eq:bernoulli}. The binary terms $x_{i}$ represents the occurrence or absence of counterfeit ROIs. Being $P( x_{i} | y )$ the likelihood of  $x_i$ given a counterfeit/genuine document $y$ and $p_{i}$ is the a priori probability of counterfeit documents in the training set. 

\begin{equation}\label{eq:bernoulli}
P(  x | y ) = \prod_{i=1}^{n} p_{i}^{x_i} (1-p_{i})^{(1-x_i)}
\end{equation}



\section{Datasets}\label{section:dataset}

ID document image datasets are restricted by copyright and government data protection rules. Hence it is not possible to obtain a public image dataset with enough resolution quality for further processing. It is also difficult to request for datasets published in counterfeit research, because most of them are embedded in a larger projects with companies, Central Banks and law enforcement authorities with non-disclosure agreements. The proposed application with the crowdsourcing mode allows to gather labeled images acquired with smartphones. These datasets can be used for incrementally train the counterfeit detection algorithms. We have created two datasets from the front and back of the Spanish ID document, see Table \ref{table:dataset}. Before implementing this framework the average time of labeling manually ESPA and ESPB images with only the cropping coordinates was $30$ seconds per image without including the acquisition time. Using the presented framework we are able to include new cropped images and process them automatically reducing the time to $12$ seconds sending $600$ dpi images, see Table \ref{table:time_results}. The reason to send the images already cropped to the server is to reduce transfer time and Internet bandwidth consumption.  

As the authors in~\cite{berenguel2016banknote} we discard all the images under $400$ dpis and resize the images ROIs to $600$ dpis, which is established as the minimum working resolution for counterfeit Intaglio feature detection~\cite{lohweg2012mobile}. Each model class ROI has a different size defined a priori by the user and the image pixel size depends on the working resolution. Setting the working resolution to $600$, we obtain class ROIs that ranges between $100 \times 100$ pixels from the smaller regions to $600 \times 600$ pixels in the bigger ROIs for the present datasets. During the whole process of dataset acquisition we alter $2$ times the image, with the format image compression of each smartphone and the warping operation in Fig \ref{grabcut}. The reason of not resizing each ROI class to the same size is to avoid introducing more artifacts to the image, which will affect in the genuine/counterfeit discrimination. Starting from the cropped document of Fig. \ref{Adjust_photo}, we can still have a shift of the texture background due the printing process of some ID documents or because the document is not perfectly aligned. Before cropping the ROI regions we perform a simple cross correlation of patches previously set for better alignment.

\begin{table}
\caption{Created datasets. Obverse (A) or reverse(R) of the document. nTextures is the number of background textures validated at each banknote/id. Train/Test is the number of documents used for train and test respectively. $\%$Counterfeit is the percentage of counterfeit samples at train/test by this order. }
\label{table:dataset}
\centering
\begin{tabular}{ccccccc}
\cline{2-5}
\hline
\multicolumn{1}{c}{Banknote}
& \multicolumn{1}{c}{nImages}
& \multicolumn{1}{c}{nTextures}
& \multicolumn{1}{c}{Train/Test}
& \multicolumn{1}{c}{\%Counterfeit} \\
\hline
\hline
		ESPA	& 1865  &	10	&	1305/560 & 	24.5/24.6	      \\ 
        ESPB	& 1268  &	7	&	887/381 & 	13.5/17.5	      \\ 
\hline
\end{tabular}%
\end{table}

\section{Experiments}\label{section:experiments}

\begin{figure}
	\centering
    \captionsetup{justification=centering}
  	\includegraphics[width=\columnwidth]{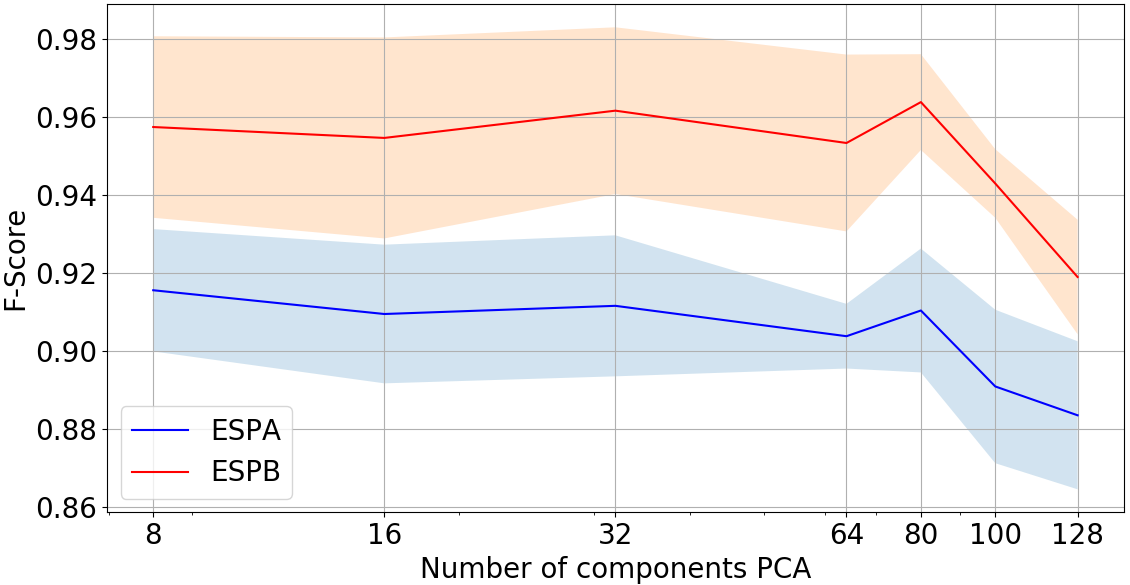}
  	\caption{Comparison between number of PCA components and $\mu$/$\sigma$ F1-score. PCA before BoVW encoder with $K=512$. X-axis in logarithmic scale. }
    \label{figure:pca}
\end{figure}

For all the experiments in the training data we set a $10$ k-fold cross-validation to optimize the SVM parameters. We repeat $5$ times a boostrapping approach to hold out a $30\%$ of the data for testing set at each dataset in all the following experiments. We repeat also $5$ times the computational time experiments but using only a test image. For the proposed benchmark all parameters are set empirically.
For feature extraction with dense SIFT we set a keypoint sampling with a step size of $s = 4$. The feature dimensionality from dense SIFT is further reduced to $D=80$ using PCA, see Fig. \ref{figure:pca}. 


\subsection{Time evaluation}

In Table \ref{table:time_results} we find a decomposition of the complete time between the acquisition and the shown results from Fig. \ref{show_results}. We include the average time for a non-experienced user from the moment they want to acquire a new document until he fits the visual guides near the document. The auto photo is the time spent to automatically acquire the photo with the best focus when the document corners are close to the visual guides. Right now, one of the bottlenecks, in time consumption, is the GrabCut and storing the image in the mobile device. However this step is needed if we want to reduce the time spent for the image transfer to the server. These times are calculated using a HSPA+ internet connection with a maximum upload speed of $5.76$ Mbits/s. Image transfer and cropped image storing time can be reduced if instead sending the full image we send and store only the ROIs needed for the evaluation. The time would depend then on the number of evaluated ROIs. So it is possible to discard the less discriminative ROIs and speed-up the process. Further work needs to be done in the mobile application to reduce the acquisition and cropping time, which actually represents approximately 93\% of the total time. The server side is processing the $10$ ROIs and returning a genuine or counterfeit document response under $1$ second. For the presented time results we use the smartphone BQ Acquaris M5 to calculate the mobile client times and the Intel(R) Xeon(R) CPU E5-1620 v3 @ 3.50GHz on the server side.

\begin{table}
\caption{ Average time in milliseconds for processing 10 ROIs of one ESPA document using BoVW with $K$=512. }
\label{table:time_results}
\centering
\begin{tabular}{cccccccc}
\hline
\multicolumn{1}{c}{}
& \multicolumn{1}{c}{}
& \multicolumn{1}{c}{$400$ dpi}
& \multicolumn{1}{c}{$600$ dpi}
& \multicolumn{1}{c}{$800$ dpi} \\
\hline
\hline
\multirow{4}{*}{Mobile app} & 		User Guide Position	& $2500$  &	$2500$	&	$2500$	 \\
        &							Auto Photo			& $980$   &	$1053$	&	$1066$	 \\
        &							GrabCut+Storing		& $4104$  &	$5930$	&	$8603$	 \\
        &							Send Image			& $1043$  &	$2113$	&	$4376$	 \\
\cline{1-2}        
\multirow{3}{*}{Server} 	&		Highlight+Blurring 	& $187$  &	$217$	&	$222$	 \\
        &							Feature Extraction 	& $389$  &	$667$	&	$861$	 \\
        &							Classification 		& $50$  &	$52$	&	$50$	 \\
\hline
\hline
		Total time			& 							& $9253$  &	$12610$	&	$17678$	 \\
\hline
\end{tabular}%
\end{table}

\begin{table*}
\caption{ $\mu$ and $\sigma$ F1-score results from different encodings with the created datasets.  }
\label{table:results}
\centering
\begin{tabular}{ccccccc}
\hline
\multicolumn{1}{c}{Banknote}
& \multicolumn{1}{c}{BoVW}
& \multicolumn{1}{c}{VLAD}
& \multicolumn{1}{c}{FV}
& \multicolumn{1}{c}{SCSPM} 
& \multicolumn{1}{c}{KSVD}	\\
\hline
\hline
		ESPA		& $0.910 \pm 0.015$  &	$0.9585 \pm 0.002$	&	\pmb{$0.981 \pm 0.005$}	&	$0.964 \pm 0.008$	&	$0.884 \pm 0.021$	      \\
        ESPB		& $0.963 \pm 0.012$  &	$0.981 \pm 0.005$	&	$0.988 \pm 0.002$	&	\pmb{$0.988 \pm 0.001$}	&	$0.939 \pm 0.018$	      \\
\hline
\end{tabular}%
\end{table*}

\subsection{Evaluation of datasets}

In table \ref{table:results} we can see the results of applying different encoders to the reduced descriptor with PCA. We set $K=512$ words for BoVW and K= $64$ for the rest of the encoders. VLAD, FV and SCSPM use a much smaller codebook as these representations multiply the dimensionality of the descriptors. K-means can be considered as a generalization of K-SVD, where only one element of the dictionary is activated each time. Relaxing the sparsity term constraint to be more than one dictionary element, K-SVD augments the representativity of the dictionary with smaller codebook.
For K-SVD we build an histogram with the absolute values of the coefficients returned from the Orthogonal Matching Pursuit algorithm (OMP). With SCSPM we partition the image into $2^{l}\times 2^{l}$ segments in different scales $l = 0,1,2$. With comparison purposes we train a PCA with $D=512$, so all the sparse coding representations are reduced to the same size. FV is among the best encoding used followed close by SCSPM and VLAD. VLAD and FV have a good performance because they encode enrich information about the visuals word's distribution. The max spatial pooling from SCSPM also prove is robust to local spatial translations. Although K-SVD results are slightly worse than BoVW, the dictionary has been reduced by a factor of $8$.

\section{Conclusion}\label{section:conclusions}
We have presented a novel application to detect ID counterfeit documents. This application is an end-to-end system that covers from the smartphone client acquisition to the evaluation of the document and final response to the client. Along the way we also build a dataset of security documents and the whole architecture schema is thought to be modular and scalable. Generating individual models for each ROI allows to introduce new documents to the system without the requirement of retraining previous models. The application can be easily extended to support banknote counterfeit detection due the strong correlation with ID background textures. One possible extension is the integration of detection and identification in the mobile application in a single step. This could be done with the recent CNN architecture presented in YOLO V2~\cite{DBLP:journals/corr/RedmonF16}. Using a joined identification and detection would facilitate the user experience and would be able to send the dewarped image once the minimum required resolution is achieved. The application will be transferred to an industrial partner to be distributed as an innovative service.





\section*{Acknowledgment}
\anonymize{
This work has been partially supported by the Spanish Research Project TIN2012-37475-C02-02 and the Industrial Doctorate Grant 2014 DI 078 with the support of the Secretariat for Universities of the Ministry of Economy and Knowledge of the Government of Catalonia.
}



%



\bibliographystyle{IEEEtran}  
\bibliography{IEEEabrv,references}

\end{document}